\documentclass[10pt,letterpaper]{article}
\usepackage[T1]{fontenc}
\usepackage[utf8]{inputenc}
\usepackage[margin=1in]{geometry}
\usepackage{times,microtype}
\usepackage{amsmath,amssymb,graphicx,booktabs,tabularx,array,xcolor,url}
\usepackage{colortbl,float}
\usepackage[round,authoryear]{natbib}
\usepackage[hidelinks]{hyperref}
\newcolumntype{Y}{>{\raggedright\arraybackslash}X}
\newcommand{\revision}[1]{#1}
\newcommand{\benchmark}{\textsc{MechBench}}
\newcommand{\domainmodel}{\textsc{MechGPT}}
\hypersetup{pdftitle={MechBench: A Domain-Specific LLM Benchmark for Parallel Mechanisms},pdfauthor={Zhihao Xia, Tingting Wang, Hui Li, Yongheng Xing, Chunxu Tian}}
\title{MechBench: A Domain-Specific LLM\\Benchmark for Parallel Mechanisms}
\author{%
  \fontsize{11}{14}\selectfont
  Zhihao Xia$^{\dagger}$\quad
  Tingting Wang$^{\dagger}$\quad
  Hui Li\quad
  Yongheng Xing\quad
  Chunxu Tian$^{*}$\\[6pt]
  \normalsize Fudan University\\[4pt]
  \small $^{\dagger}$Co-first authors (equal contribution).\quad $^{*}$Corresponding author.
}

\date{}
\begin{document}
\maketitle

\begin{abstract}
\revision{Parallel mechanisms pose distinctive challenges to large language models (LLMs), as valid reasoning depends jointly on mechanism topology, loop closure, geometry, joint types, and modeling assumptions. Yet existing LLM benchmarks provide little systematic coverage of such domain-specific knowledge, constraint-sensitive analysis, and constructive design. To fill this gap, we introduce \textsc{MECHBENCH}, a 2,541-question benchmark for evaluating LLMs in parallel mechanisms across professional knowledge, mechanism analysis, and design. \textsc{MECHBENCH} contains standard and challenge tiers spanning six capability categories and nine executable analysis and design tasks. It is constructed through \textsc{MECHFORGE}, a domain-grounded framework that combines expert-organized knowledge, LLM-assisted question generation, programmatic instance generation, two-stage expert review, and task-aware scoring with exact matching, rubric-based evaluation, and program verification. We further develop \textsc{MECHGPT} as a diagnostic domain-adapted model. Experiments show that strong knowledge performance does not directly translate into executable mechanism reasoning: leading closed-service models achieve over 95 on the standard tier but only 47--60 on the challenge tier, while domain adaptation substantially improves MechGPT over its Qwen3-8B baselines but leaves a large gap in constrained analysis and design. These results establish \textsc{MECHBENCH} as a systematic testbed for measuring domain-specific LLM capabilities in mechanism science.}

\end{abstract}

\section{Introduction}

\revision{Parallel-mechanism reasoning depends jointly on topology, loop closure, geometry, joint types, and modeling assumptions. A valid evaluation must test whether an LLM preserves these conditions, selects an appropriate method, and reaches a compatible conclusion. General benchmarks such as MMLU, GPQA, MMLU-Pro, and Humanity's Last Exam provide broad knowledge tests but do not isolate these requirements \citep{mmlu,gpqa,mmlupro,hle}.}

\revision{The domain presents specific analytical distinctions. Closed-loop singularities and Jacobian-based performance measures depend on configuration and interpretation \citep{gosselin1990singularity,merlet2006}; configurable platforms introduce internal motion that differs from output mobility \citep{mohamed2005configurable,lambert2016configurable}. Generalized mechanisms further involve coupled sub-chains and reconfigurable working phases \citep{xia2025gpmreview,tian2024redundancy}. Evidence is distributed across papers, textbooks, exercises, and code with differing symbols and assumptions, making direct QA collection vulnerable to incomplete conditions and uneven coverage.}

\revision{To address these challenges, we propose \textsc{MechForge}, a framework for constructing domain-specific benchmarks and instruction datasets from professional materials (Figure~\ref{fig:framework}). Building on knowledge-organized data construction \citep{oceangpt,guo2025sdbench}, it combines LLM-assisted architecture expansion and evidence-grounded QA generation with programmatic instances, model checking, and two-stage expert review. The checks preserve connectivity, geometry, constraint compatibility, and the distinction between whole-mechanism mobility, output motion, and internal motion.}

\revision{MechBench is the central evaluation artifact, designed to distinguish domain knowledge from constraint-sensitive mechanism analysis and constructive design; MechForge specifies its construction methodology, and MechGPT serves as a diagnostic instance for studying capability changes after domain adaptation. The contributions are:}
\begin{itemize}
\item \revision{We introduce \textbf{MechBench}, a 2,541-question benchmark for parallel mechanisms that separates domain knowledge from condition-sensitive analysis and constructive design, covering topology, mobility, modeling, diagnosis, mechanism modification, completion, and constrained synthesis.}
\item \revision{We propose \textbf{MechForge}, a domain-grounded construction methodology combining knowledge organization, LLM-assisted generation, two-stage expert review, and executable checks of topology, geometry, constraints, and distinct mobility concepts.}
\item \revision{We evaluate closed-service and open-weight models, including the domain-adapted \textbf{MechGPT}, using mean scores and capability/task profiles. The results expose a clear gap between high professional-knowledge performance and executable, constraint-consistent mechanism reasoning and design.}
\end{itemize}

\section{Related Work}

\paragraph{LLM Benchmarking.}
\revision{General benchmarks assess broad knowledge and expert reasoning \citep{mmlu,gpqa,mmlupro,hle}. SciKnowEval organizes scientific capabilities by cognitive operation \citep{sciknoweval}, while SciR generates verifiable questions from formalized objects and scientific text \citep{scir}. Open-ended scoring introduces judge biases \citep{mtbench}; G-Eval and Prometheus 2 address evaluation through structured steps and explicit rubrics \citep{geval,prometheus2}. MechBench combines exact matching, program verification, and item-specific rubrics, with recorded scoring configurations and sampled human validation.}

\paragraph{Domain-specific LLM and Data Construction.}
\revision{Retrieval supplies external evidence at inference time \citep{lewis2020rag}; continued pretraining and instruction tuning incorporate domain knowledge into model parameters \citep{dontstop,llemma,pmcllama,adaptllm}. LoRA and GaLore reduce adaptation costs \citep{hu2021lora,galore}. Self-Instruct and Bonito generate instruction data from text \citep{selfinstruct,bonito}, while OceanGPT and SDBench link knowledge organization to training and evaluation \citep{oceangpt,guo2025sdbench}. Engineering benchmarks require document-grounded constraints \citep{designqa}, complementing executable robotics interfaces such as Code as Policies \citep{codepolicies}. MechBench focuses on condition-sensitive mechanism reasoning and records source relationships for overlap and contamination auditing \citep{contamination}.}

\paragraph{\revision{Parallel-Mechanism Analysis and Synthesis.}}
\revision{Classical studies analyze closed-loop singularities \citep{gosselin1990singularity}, Jacobian-based performance \citep{merlet2006}, and configurable-platform topology, mobility, and overconstraints \citep{mohamed2005configurable,lambert2016configurable}. Later work combines graph connectivity with screw constraints for coupled sub-chains \citep{tian2017coupling}, matches limb constraints to platform motion \citep{tian2020configurable}, and studies reconfigurable working phases \citep{tian2021reconfigurable}. Motion-transmission synthesis and topological mapping networks connect structure to desired outputs \citep{xia2025transmission,xia2026mapping}. These distinctions guide the structural and analytical checks in MechBench.}

\section{\revision{MechForge Construction Framework}}
\label{sec:framework}

\begin{figure}[t]
    \centering
    \includegraphics[width=\linewidth]{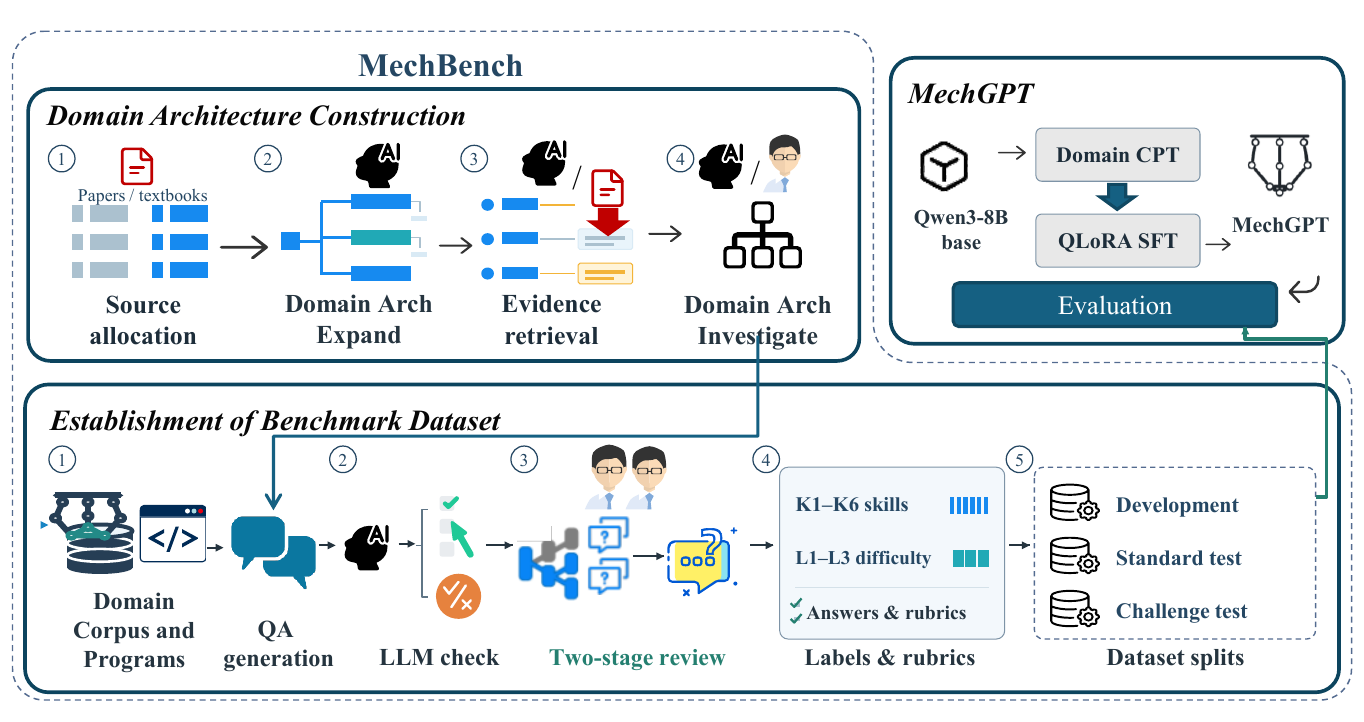}
    \caption{\revision{MechForge workflow. Sources feed LLM-assisted knowledge organization and evidence-grounded or programmatic question generation. LLM checking and two-stage researcher review confirm conditions, answers, labels, and rubrics before tier assignment; reviewed QA can also form instruction data. Domain pretraining and QLoRA adapt MechGPT, which is evaluated by MechBench.}}
    \label{fig:framework}
\end{figure}

\revision{Figure~\ref{fig:framework} organizes MechForge into knowledge-architecture construction and benchmark establishment, with a separate branch connecting the resulting resources to model adaptation and evaluation.}

\subsection{Knowledge Architecture and Source Organization}
\label{sec:knowledge_architecture}

\begin{enumerate}\setlength{\itemsep}{1pt}
    \item \revision{\textbf{Source Allocation.} Organize papers, surveys, textbooks, teaching materials, and researcher-written generators by use in training, knowledge organization, and question generation; record source relationships.}
    \item \revision{\textbf{Domain Architecture Expansion.} An LLM proposes topics spanning topology, geometry, constraints, mobility, singularities, synthesis, and diagnosis, distinguishing rigid platforms from coupled, configurable, or articulated architectures.}
    \item \revision{\textbf{Evidence Retrieval.} Retrieve and prioritize supporting passages, retaining joint-axis geometry, platform assumptions, and working phases that determine each relation's applicability.}
    \item \revision{\textbf{Domain Architecture Investigation.} Researchers check the proposed architecture against evidence, refine topic boundaries, and assess coverage. Technical topics and K1--K6 capability labels remain distinct.}
\end{enumerate}

\subsection{Establishment of the Benchmark Dataset}
\label{sec:benchmark_construction}

\revision{An LLM receives a topic, prioritized source evidence, and response-format requirements to generate questions, proposed answers, and scoring points. Programmatic generators supply complementary instances with controlled answers. Both paths follow the same review sequence:}
\begin{enumerate}\setlength{\itemsep}{1pt}
    \item \revision{\textbf{QA Generation.} Generate choice, fill-in, or constructed-response items. State the required base, platform, limb connectivity, joint types, geometry, and actuation; retain shared links and joints in coupled sub-chains.}
    \item \revision{\textbf{LLM Check.} A second LLM checks evidence, condition sufficiency, answer consistency, and format. Mechanism checks cover loop closure, constraint independence, platform-motion compatibility, and applicable input--output relations.}
    \item \revision{\textbf{Two-Stage Expert Review.} Two researchers sequentially review every item, resolve disagreements, and revise or reject invalid candidates. They distinguish whole-mechanism mobility, output and internal motion, instantaneous and finite motion, and reconfiguration phases.}
    \item \revision{\textbf{Labels and Rubrics.} Confirm K1--K6 capabilities, L1--L4 difficulty, reference answers, equivalent expressions, and scoring points. Intentionally under-specified items must identify the missing condition and its consequence.}
    \item \revision{\textbf{Dataset Splits.} Assign accepted items to development, standard, or challenge tiers; retain source-group identifiers for overlap auditing, and validate fields, answer keys, legal scores, and coverage.}
\end{enumerate}

\section{Case Study: \benchmark{} and \domainmodel}
\label{sec:case_study}

\revision{We apply MechForge to parallel mechanisms and connect the resulting benchmark to MechGPT. Adaptation starts from Qwen3-8B-Base through domain continued pretraining followed by supervised fine-tuning. Main evaluation covers the 1,791-question standard tier and nine tasks in the 550-question challenge tier; development questions are used only for protocol debugging.}

\subsection{Domain Corpus}

\revision{We assemble 10,379 professional papers, textbooks, and teaching materials. The main continued-pretraining collection contains 9,869 documents, including 93 textbooks and supplementary exercises, with professional papers constituting the remainder. Conversion to unified text removes author information, headers, footers, references, empty lines, and extraction noise while retaining technical relations and their assumptions.}

\revision{Instruction-data sources comprise 202 English papers, 28 deduplicated textbooks, and 12 sets of course materials; the papers yield 56,927 QA pairs. Researcher-written examples, QA transformation, and model-assisted generation followed by cleaning supply domain dialogues. Training records use a unified conversational representation and the Qwen3 chat template.}

\subsection{Parallel-Mechanism Knowledge Architecture}

\begin{figure}[t]
    \centering
    \includegraphics[width=0.52\linewidth]{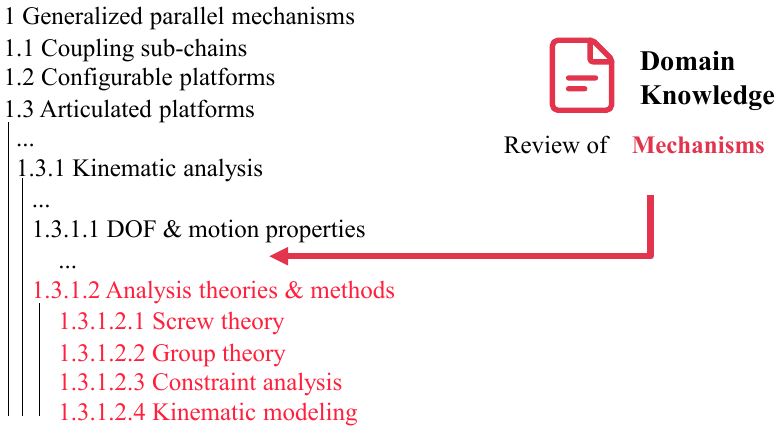}
    \caption{\revision{Knowledge branch linking coupled and configurable architectures to motion properties and screw-, group-, constraint-, and model-based analysis.}}
    \label{fig:knowledge_architecture}
\end{figure}

\revision{Figure~\ref{fig:knowledge_architecture} illustrates the topic hierarchy. Capability, difficulty, response format, and scoring path are labeled independently. Six capabilities distinguish \textbf{K1}, terms and symbols; \textbf{K2}, structure and topology; \textbf{K3}, mobility and kinematics; \textbf{K4}, modeling, calibration, and analysis; \textbf{K5}, diagnosis and information sufficiency; and \textbf{K6}, synthesis and design trade-offs. K3 asks for a motion or quantity; K4 asks how to model or obtain it. Expert-assigned difficulty spans L1 (recognition or one-step application) to L4 (complex synthesis, proof, or path verification), independently of model accuracy.}

\subsection{\benchmark}
\label{sec:benchmark}

\revision{Each item has an identifier, stem, required response format, and applicable choices or fill-in fields. Separately managed scoring resources contain reference answers, acceptable alternatives, scoring points, and error categories. Constructed responses are explanations, formulas, or derivations.}

\revision{MechBench contains 200 development, 1,791 standard, and 550 challenge questions (2,541 total; Table~\ref{tab:dataset_distribution}). The development tier supports protocol debugging. The original 1,991 items comprise development and standard tiers; the separate 550-item set is the challenge tier.}

\begin{table}[t]
\centering
\color{black}
\caption{Composition of \benchmark.}
\label{tab:dataset_distribution}
\small
\begin{minipage}[t]{0.48\linewidth}
\vspace{0pt}
\centering
\setlength{\tabcolsep}{3pt}
\renewcommand{\arraystretch}{1.25}
\begin{tabularx}{\linewidth}{@{}Yrr@{}}
\toprule
\multicolumn{3}{c}{\textbf{(a) Benchmark tiers}}\\
\multicolumn{3}{c}{All questions: $N=2,541$}\\
Tier & Questions & Share (\%)\\
\midrule
Development & 200 & 7.87\\[5pt]
Standard & 1,791 & 70.48\\[5pt]
Challenge & 550 & 21.65\\[5pt]
Total & 2,541 & 100.00\\
\bottomrule
\end{tabularx}
\end{minipage}\hfill
\begin{minipage}[t]{0.48\linewidth}
\vspace{0pt}
\centering
\setlength{\tabcolsep}{3pt}
\renewcommand{\arraystretch}{1.25}
\begin{tabularx}{\linewidth}{@{}lYr@{}}
\toprule
\multicolumn{3}{c}{\textbf{(b) Professional capabilities}}\\
\multicolumn{3}{c}{Standard questions: $N=1,791$}\\
 & Capability & Share (\%)\\
\midrule
K1 & Terms/Symbols & 10.94\\
K2 & Structure/Topology & 14.41\\
K3 & Mobility/Kinematics & 44.11\\
K4 & Modeling/Analysis & 17.87\\
K5 & Diagnosis/Sufficiency & 7.43\\
K6 & Design Trade-offs & 5.25\\
\bottomrule
\end{tabularx}
\end{minipage}
\end{table}

\begin{figure}[t]
\centering
\includegraphics[width=0.84\linewidth]{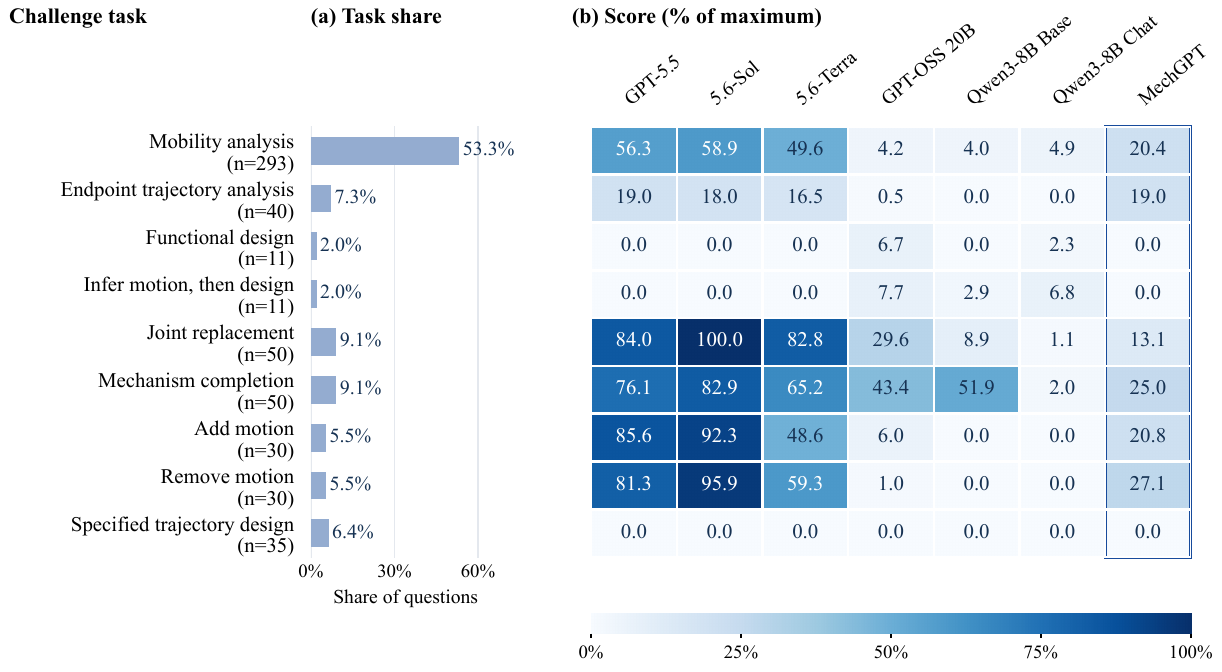}
\caption{\revision{Challenge-task shares and normalized means; Qwen3-8B Base and Qwen3-8B Chat are shown with their full model names.}}
\label{fig:challenge_profile}
\end{figure}

\revision{Standard questions cover six capabilities; K3 and K4 together account for 61.98\%. The challenge tier covers nine analysis and design tasks (Figure~\ref{fig:challenge_profile}). Tier membership is distinct from L1--L4 difficulty. Source-group identifiers support overlap auditing.}

\begin{figure*}[t]
    \centering
    \includegraphics[width=0.86\textwidth]{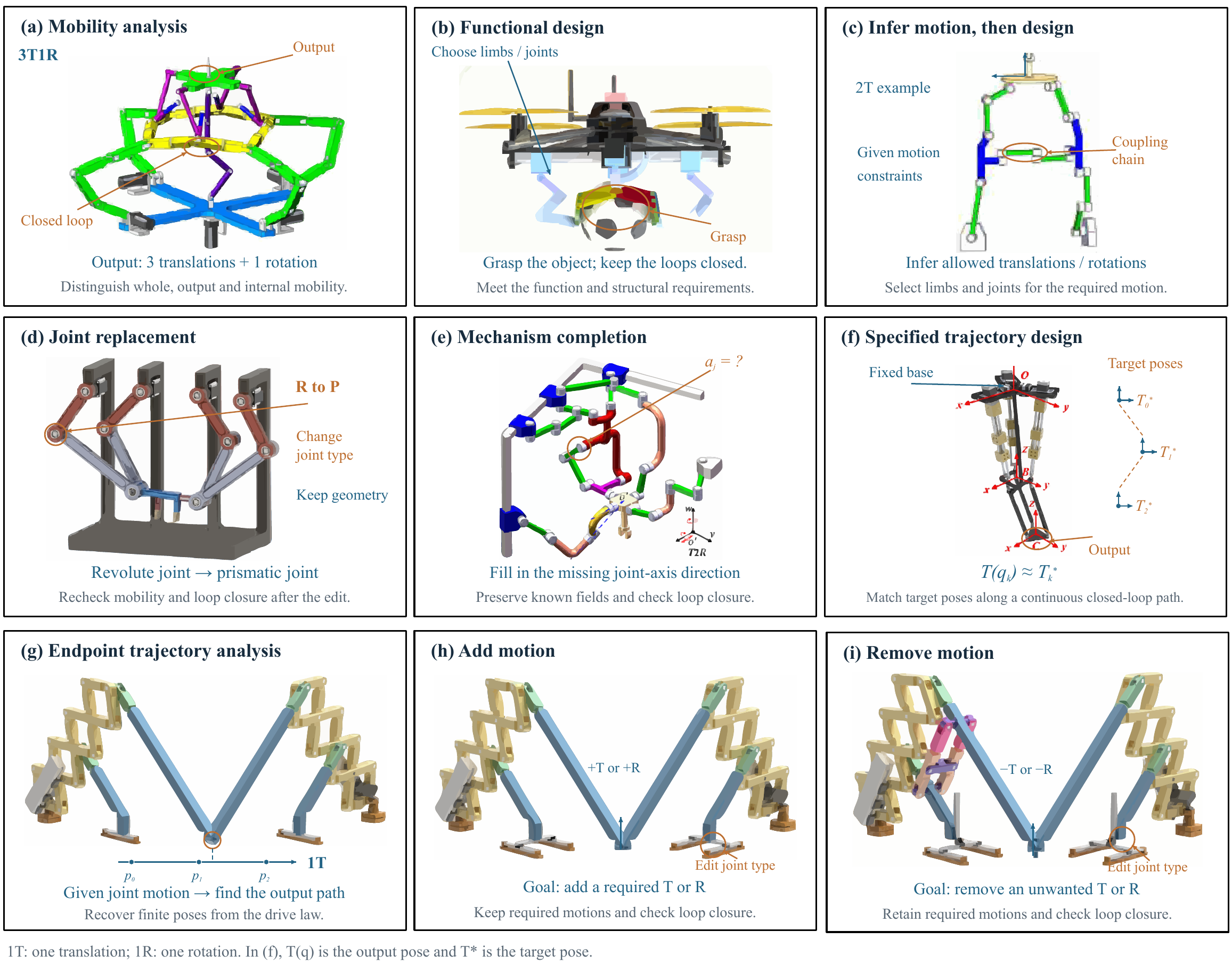}
    \caption{\revision{Mechanism illustrations for the nine challenge tasks, spanning analysis, modification, completion, and constrained design.}}
    \label{fig:challenge_task_examples}
\end{figure*}

\revision{Figure~\ref{fig:challenge_task_examples} illustrates the mechanism content assessed by the nine challenge tasks.}

\revision{Table~\ref{tab:question_examples} summarizes question types and representative stems. Analysis and design group the nine challenge tasks.}

\begin{table}[t]
\centering\color{black}
\caption{Question formats, counts, and abbreviated example stems.}
\label{tab:question_examples}
\small
\setlength{\tabcolsep}{4pt}
\renewcommand{\arraystretch}{1.08}
\begin{tabularx}{\linewidth}{@{}>{\raggedright\arraybackslash}p{0.17\linewidth}rY@{}}
\toprule
Type & Count & Example question\\
\midrule
Single-choice & 870 & Before loop closure, a parallel mechanism has 10 independent joint-motion parameters; closure makes 7 dependent. What is its mobility? A. 3; B. 7; C. 10; D. 17.\\
\addlinespace[2pt]
Multiple-choice & 466 & Which methods derive the platform-velocity Jacobian? A. Loop equations; B. Reciprocal screws; C. Pl\"ucker coordinates; D. Limb counting alone.\\
\addlinespace[2pt]
Fill-in & 358 & An ideal spherical joint permits relative rotations about its center. It has \underline{\hspace{1.5em}} relative rotational degrees of freedom.\\
\addlinespace[2pt]
Constructed response & 297 & What defines a general screw? Which factors remain for a normalized unit screw?\\
\addlinespace[2pt]
Analysis & 333 & Three UP limbs with disjoint links connect the base and platform\ldots{} Report first- and second-order mobility, output and self-motion mobility, and an output-twist basis.\\
\addlinespace[2pt]
Design & 217 & A platform must satisfy $C\xi=0$ without extra locking\ldots{} Infer its motion and design a four-limb mechanism with geometry and motion-screw justification.\\
\addlinespace[2pt]
\textbf{Total} & \textbf{2,541} & \\
\bottomrule
\end{tabularx}
\end{table}

\subsection{Task-Aware Scoring}
\label{sec:scoring}

\begin{figure}[t]
    \centering
    \includegraphics[width=\linewidth]{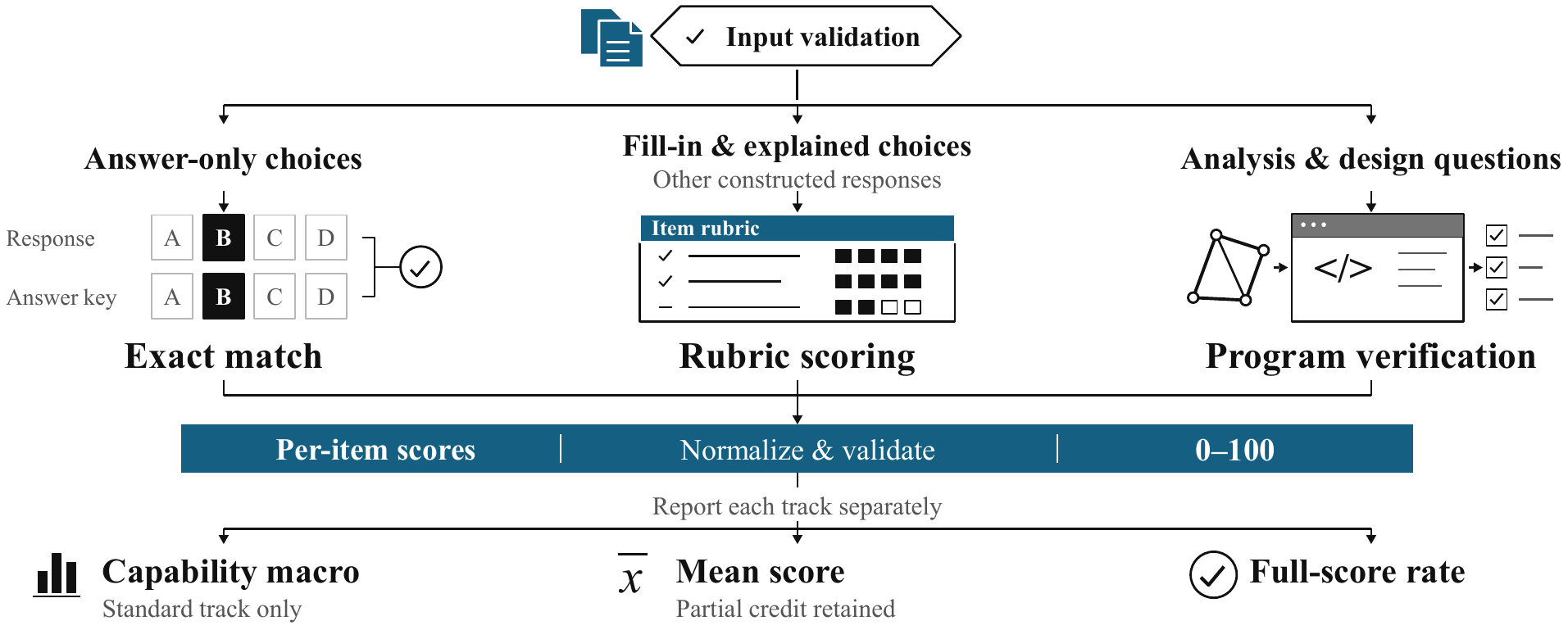}
    \caption{\revision{Scoring routes: exact matching, item rubrics, or program checks, with scores normalized to 0--100.}}
    \label{fig:scoring}
\end{figure}

\revision{Development and standard answer-only choices use exact matching; explained choices, fill-in questions, and constructed responses use item-specific 0--5 rubrics normalized to 0--100. Challenge questions use deterministic verification on a 0--100 scale. Both assess correctness under stated structural, geometric, and motion conditions. Figure~\ref{fig:scoring} shows the routing and normalization steps.}

\revision{Let $I_m$ contain numerically scored items and $s_{mi}\in[0,100]$. We report the normalized mean score (micro-average), and the equal-weight mean over capability or task groups $G$:}
\begin{equation}
\color{black} S_m=\frac{\sum_{i\in I_m}s_{mi}}{|I_m|},\qquad M_m=\frac{1}{|G|}\sum_{g\in G}\frac{\sum_{i\in I_m\cap g}s_{mi}}{|I_m\cap g|}.
\label{eq:score}
\end{equation}
\revision{Standard capability macro uses six groups and challenge task macro uses nine. Micro means remain primary; macro scores provide a distribution-balanced supplement. Unresolved execution failures have no numerical score, whereas invalid submitted structures receive zero.}

\paragraph{\revision{Executable mechanism checks.}}
\revision{Mechanisms encode links, a fixed base, an output platform, and R/P/U/S joints with incident links, positions, and axes. Analysis answers distinguish first-order candidate mobility, second-order-compatible mobility, output mobility, and internal self-motion; design answers provide a mechanism payload. MechOmni-based checks assess schema validity, closure, limb and joint constraints, output screw spaces, and applicable finite-motion or trajectory conditions. Screw-space comparison accepts equivalent bases rather than requiring identical vector order, and rubric-weighted checks award partial credit while retaining designated design hard gates.}

\noindent\begin{minipage}{\linewidth}
\paragraph{\revision{Algorithm 2: Task-Aware Item Scoring.}}
\begin{quote}\small\color{black}
\textbf{Input:} Response, reference, and item configuration. \textbf{Output:} Score and evidence. Validate identity; route to matching, the recorded rubric-judge configuration, or program checks; adjudicate where configured; retain criterion evidence and execution status; normalize and aggregate by tier, capability, and task. Faithful extraction may recover explicit fields but may not solve, correct, or complete the answer.
\end{quote}
\end{minipage}

\section{Experiment}

\subsection{Implementation Details}

\paragraph{Implementation of \benchmark.}
\revision{LLM- and program-generated candidates undergo a full first researcher review and a second-stage full review, covering conditions, answers, labels, and rubrics. The second stage is not claimed to be blinded or statistically independent. Field validation is separate from professional-content review.}

\paragraph{Implementation of \domainmodel.}
\revision{We select Qwen3-8B-Base as the foundation model \citep{yang2025qwen3}. MechGPT first undergoes continued pretraining on the collected mechanism-domain corpus, followed by supervised fine-tuning. Training uses eight NVIDIA GeForce RTX 4090 GPUs with 24 GB of memory each. Continued pretraining runs for one epoch with a learning rate of $1\times10^{-6}$. Supervised fine-tuning uses 4-bit QLoRA with a learning rate of $1\times10^{-5}$ for three epochs.}

\paragraph{\revision{Compared models.}}
\revision{Both collections compare GPT-5.5, GPT-5.6-Sol, GPT-5.6-Terra, GPT-OSS 20B, Qwen3-8B Base and Chat, and MechGPT.}

\paragraph{\revision{Evaluation protocol.}}
\revision{The standard tier combines answer matching and rubric scoring under a unified evaluation configuration; challenge tasks use program verification.}

\subsection{\revision{Benchmark Composition and Quality}}
\revision{Development questions support protocol debugging and are excluded from main results. Standard questions cover six capabilities; challenge questions cover nine tasks. Construction review confirms each item's capability and difficulty labels, evidence, conditions, reference answers, and scoring operability.}

\revision{To reduce training--evaluation contamination, we manually separated training and Standard/Challenge materials by source group during curation. An original paper, textbook chapter, code document, or generation dependency and its rewrites, QA pairs, program transformations, and generated variants were assigned to the same group. Standard/Challenge source groups were withheld from training; splitting only by question identifier was not used.}

\revision{Scoring follows two validation routes. Objective and program-scored questions are processed by fixed answer rules or deterministic mechanism checks. For open-ended questions scored with item-specific rubrics, domain experts reviewed a sample of the model-based judgments and confirmed their consistency with the reference answers and scoring criteria. For this calibration check, every model was compared on the same 150 questions: 50 multiple-choice, 50 fill-in, and 50 open-ended items. Table~\ref{tab:human_recheck} reports the historical and expert-reviewed aggregate means; the comparison is descriptive and does not replace item-level agreement analysis.}

\begin{table}[t]
\centering
\scriptsize
\setlength{\tabcolsep}{3.2pt}
\caption{Expert recheck on the same 150 questions per model (50 per format); $\Delta$ is expert minus historical mean.}
\label{tab:human_recheck}
\begin{tabular}{@{}lrrrrrr@{}}
\toprule
Model & Historical & Expert & $\Delta$ & Multiple-choice & Fill-in & Open-ended \\
\midrule
GPT-5.5 & 93.07 & 92.43 & $-0.64$ & 98.40 & 82.20 & 96.70 \\
GPT-5.6-Sol & 94.80 & 91.07 & $-3.73$ & 96.40 & 79.00 & 97.80 \\
GPT-5.6-Terra & 93.60 & 90.47 & $-3.13$ & 98.00 & 78.20 & 95.20 \\
GPT-OSS-20B & 73.20 & 70.43 & $-2.77$ & 82.00 & 56.20 & 73.10 \\
Qwen3-8B Chat & 61.06 & 57.37 & $-3.69$ & 62.00 & 44.40 & 65.70 \\
Qwen3-8B Base & 42.94 & 43.13 & $+0.19$ & 69.60 & 19.00 & 40.80 \\
MechGPT & 62.53 & 63.97 & $+1.44$ & 79.60 & 51.60 & 60.70 \\
\bottomrule
\end{tabular}
\end{table}

\subsection{\revision{Benchmark Results and Analysis}}
\label{sec:experiment_results}

\paragraph{\revision{Knowledge performance does not directly translate to executable mechanism reasoning.}}
\revision{GPT-5.6-Sol achieves the highest scores on both the standard and challenge tiers, with scores of 96.12 and 59.68, respectively. MechGPT obtains 62.35 on the standard tier but only 18.30 on the challenge tier, revealing a substantial gap between conventional domain knowledge and executable mechanism reasoning.}

\revision{On the challenge tier, Sol, GPT-5.5, and Terra score 59.68, 55.15, and 46.95. MechGPT scores 18.30, compared with GPT-OSS 20B's 8.86, Base's 7.36, and Chat's 3.25. Chat's advantage over Base on standard questions reverses on the challenge tier, showing that stronger knowledge-oriented performance does not ensure reliable structural analysis or design.}

\begin{figure}[H]
\centering
\includegraphics[width=0.82\linewidth]{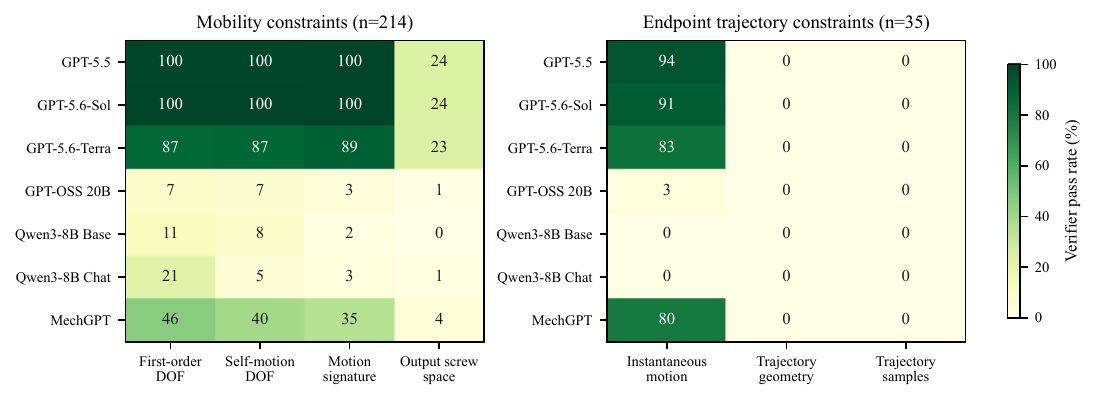}
\caption{\revision{Verifier pass rates on paired mobility ($n=214$) and endpoint-trajectory ($n=35$) questions.}}
\label{fig:constraint_diagnostic}
\end{figure}

\paragraph{\revision{Constraint-level diagnostic.}} \revision{Figure~\ref{fig:constraint_diagnostic} summarizes the verifier-level pass rates.}
\revision{To explain the gap between aggregate challenge scores, we inspect verifier criteria on paired subsets. On 214 mobility questions with complete evidence for all models, GPT-5.6-Sol passes all scalar mobility and motion-signature checks, but its strict output screw-space pass rate is 23.8\%. MechGPT reaches 46.3\%, 39.7\%, and 35.0\% on first-order mobility, self-motion, and motion-signature checks, respectively, but 4.2\% on output screw space. These results indicate that recognizing a motion class is easier than constructing a geometry-consistent six-dimensional motion basis. On 35 paired endpoint-trajectory questions, instantaneous-motion pass rates remain high for the strongest models, whereas trajectory geometry and trajectory samples receive no passes in the current records.}

\begin{table}[t]
\centering
\scriptsize
\setlength{\tabcolsep}{2.7pt}
\caption{Standard capability means (0--100; $n=1,791$). Base/Chat denote Qwen3-8B. GPT-OSS 20B values use the supplied capability-level means.}
\label{tab:ability_scores}
\begin{tabular}{@{}lrrrrrrr@{}}
\toprule
Capability & GPT-5.5 & Sol & Terra & GPT-OSS 20B & Base & Chat & MechGPT \\
\midrule
K1 (n=196) & 88.6 & 88.4 & 87.0 & 69.79 & 41.1 & 51.8 & 69.9 \\
K2 (n=258) & 93.7 & 95.3 & 94.9 & 83.58 & 43.5 & 50.2 & 68.8 \\
K3 (n=790) & 97.6 & 97.3 & 96.9 & 81.05 & 39.8 & 45.4 & 59.0 \\
K4 (n=320) & 96.7 & 96.9 & 94.6 & 85.99 & 38.7 & 49.8 & 61.8 \\
K5 (n=133) & 98.2 & 98.6 & 98.9 & 87.33 & 54.4 & 66.4 & 61.1 \\
K6 (n=94) & 94.9 & 98.5 & 96.2 & 71.49 & 40.3 & 53.0 & 60.6 \\
\bottomrule
\end{tabular}
\end{table}

\revision{Table~\ref{tab:ability_scores} compares capability profiles on the standard tier. MechGPT exceeds Base in all six capabilities and Chat in five. Gains over Chat are largest for structure/topology (K2, +18.57) and terms/symbols (K1, +18.06); diagnosis/sufficiency (K5) remains weaker, at 61.05 versus 66.43.}

\revision{On the challenge tier, mobility analysis occupies 53.27\% of questions. MechGPT scores 20.39 versus Base's 4.03 and Chat's 4.91, but below Sol's 58.86. It exceeds both Qwen baselines on mobility, endpoint trajectory analysis, joint replacement, adding motion, and removing motion. Mechanism completion reverses that pattern: 25.02 versus Base's 51.90. Endpoint analysis reaches only 19.00, and no evaluated model obtains credit for specified trajectory design.}

\paragraph{\revision{Feasibility sanity check.}} \revision{Zero scores on the 35 specified-trajectory designs do not imply infeasibility: all references pass the checks in Table~\ref{tab:trajectory_feasibility}, and ten independently selected instances admit full-score executable constructions. Perturbing axis direction, closure geometry, joint type, or target trajectory is rejected or capped. The difficulty lies in jointly synthesizing valid topology, closure, target mobility, and a continuous trajectory certificate.}

\begin{table}[t]
\centering
\scriptsize
\setlength{\tabcolsep}{4pt}
\caption{Feasibility checks for specified-trajectory design references.}
\label{tab:trajectory_feasibility}
\begin{tabular}{@{}lrlr@{}}
\toprule
Check & Passed & Check & Passed \\
\midrule
Payload parsing & 35/35 & Target mobility/output & 35/35 \\
Topology constraints & 35/35 & Finite continuation & 35/35 \\
Initial closure & 35/35 & Pose trajectory & 35/35 \\
Independent constructions & 10/10 & & \\
\bottomrule
\end{tabular}
\end{table}

\subsection{Future Work}

\revision{Based on feedback from parallel-mechanism researchers during construction and internal evaluation, we will refine MechForge and MechGPT in two directions. First, we will adopt stronger open-weight foundations to improve deeper reasoning and analysis. Second, we will extend both systems to mechanism drawings, CAD and geometric representations, simulation trajectories, and experimental sensor signals.}

\section{Conclusion}

\revision{In this paper, we introduce MechBench, a benchmark for evaluating professional knowledge, condition-sensitive analysis, and constructive design in parallel mechanisms. MechBench contains 2,541 questions organized into development, standard, and challenge tiers and combines exact matching, rubric scoring, and program verification. Its standard and challenge tracks reveal a substantial gap between conventional domain knowledge and executable, constraint-consistent mechanism reasoning. MechForge provides the construction methodology behind the benchmark, while MechGPT serves as a domain-adapted diagnostic model.}

\section{Limitation}

\revision{MechForge organizes domain evidence and structural conditions into a benchmark that covers knowledge comprehension, mechanism analysis, and design. However, the framework does not yet automatically control the quality of the corpus used for continued pretraining. Selecting reliable sources and checking the adequacy of mechanism descriptions still require expert judgment. In addition, MechBench currently evaluates text and structured mechanism representations rather than multimodal inputs. This limits its coverage of engineering problems that require interpreting mechanism drawings, spatial geometry, and physical interaction.}

\revision{For MechGPT, domain adaptation improves knowledge and analysis but does not consistently extend to completion or constrained synthesis. Future versions will add geometric representations and training examples with explicit constraint checks.}

\clearpage
\appendix
\onecolumn
\section{Challenge Response Visualization}
\begin{figure}[H]
\centering
\includegraphics[width=0.90\textwidth,height=0.40\textheight,keepaspectratio]{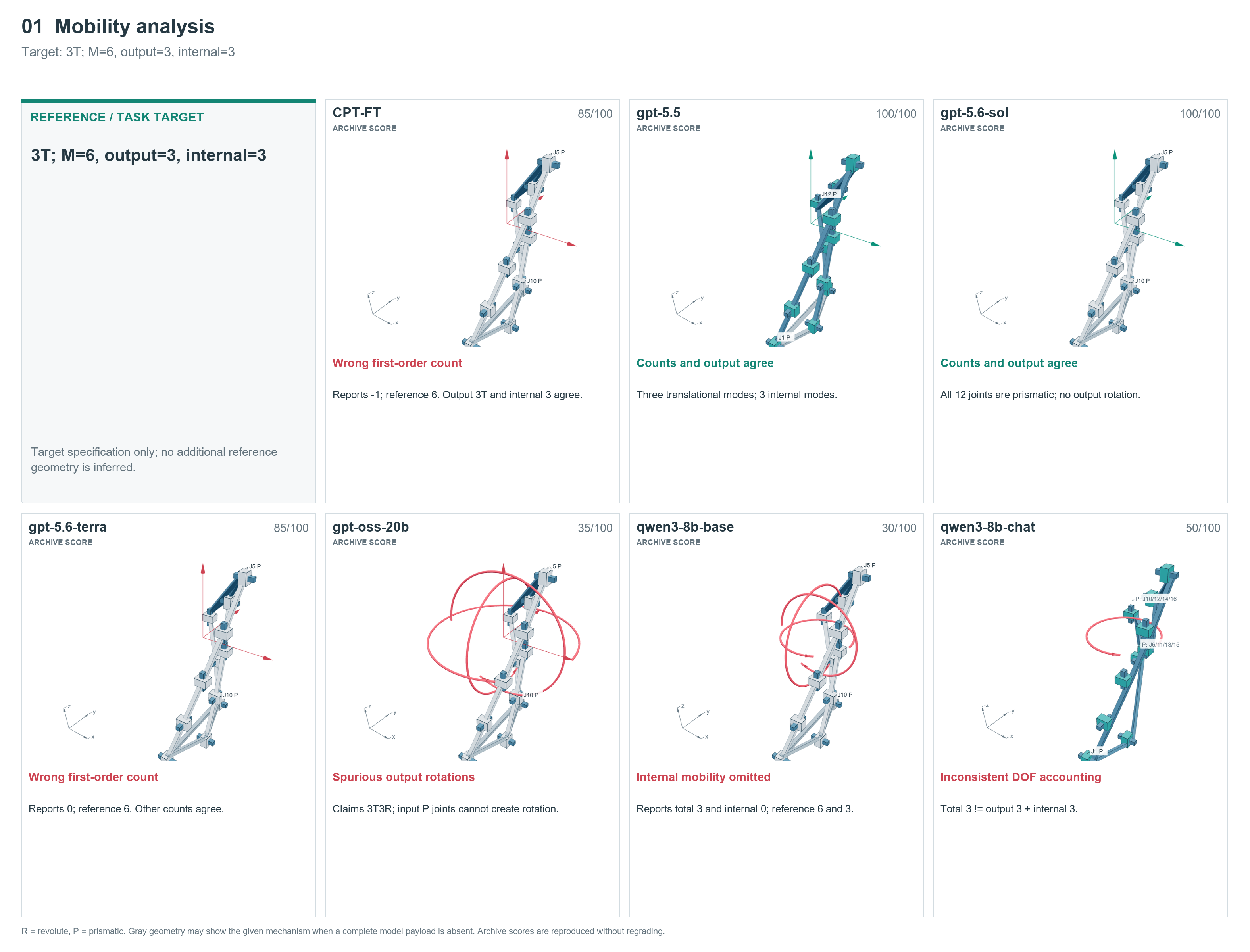}\\[4pt]
\includegraphics[width=0.90\textwidth,height=0.40\textheight,keepaspectratio]{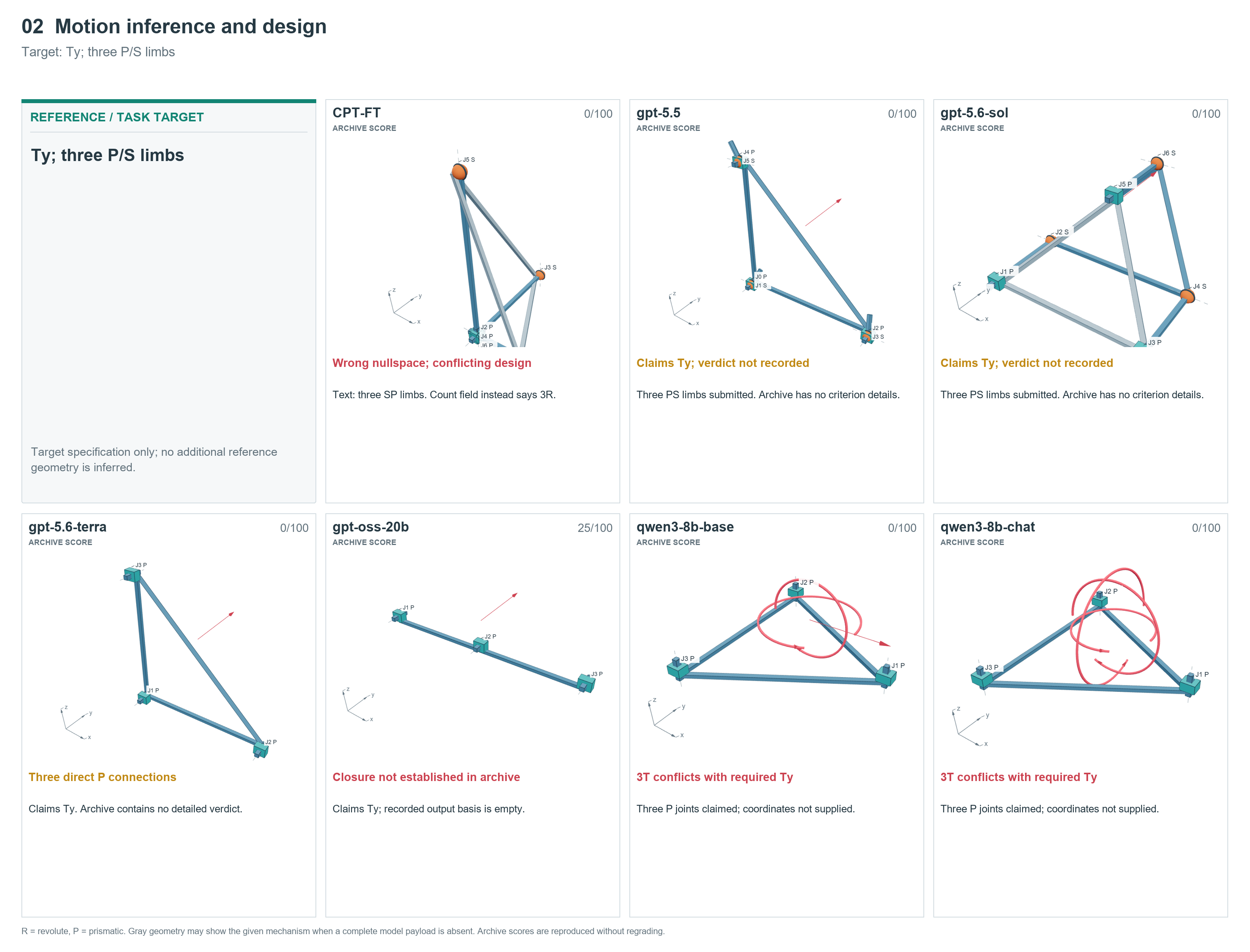}
\caption{Challenge response visualizations for categories 1--2.}
\end{figure}
\clearpage
\begin{figure}[H]
\centering
\includegraphics[width=0.90\textwidth,height=0.40\textheight,keepaspectratio]{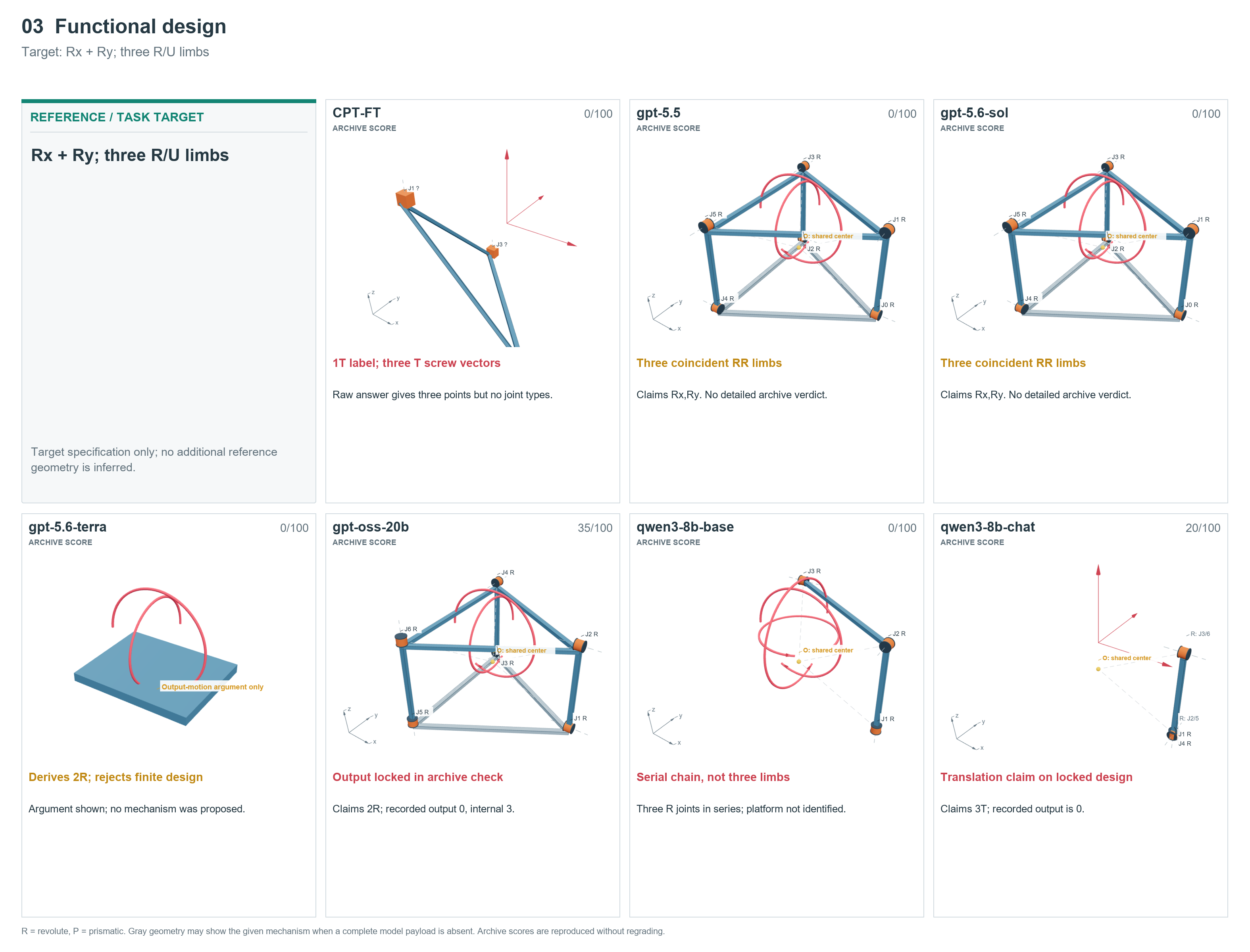}\\[4pt]
\includegraphics[width=0.90\textwidth,height=0.40\textheight,keepaspectratio]{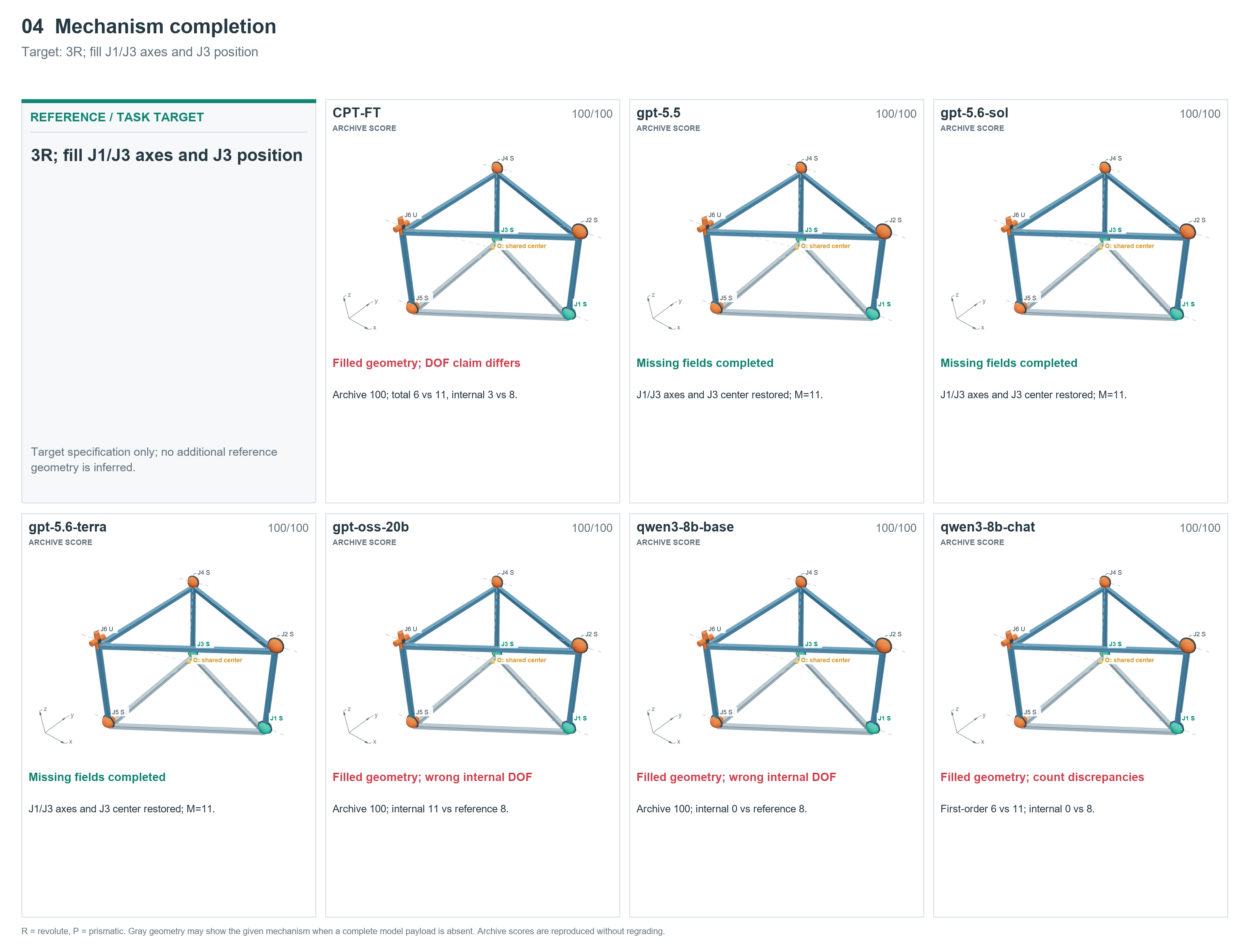}
\caption{Challenge response visualizations for categories 3--4.}
\end{figure}
\clearpage
\begin{figure}[H]
\centering
\includegraphics[width=0.90\textwidth,height=0.40\textheight,keepaspectratio]{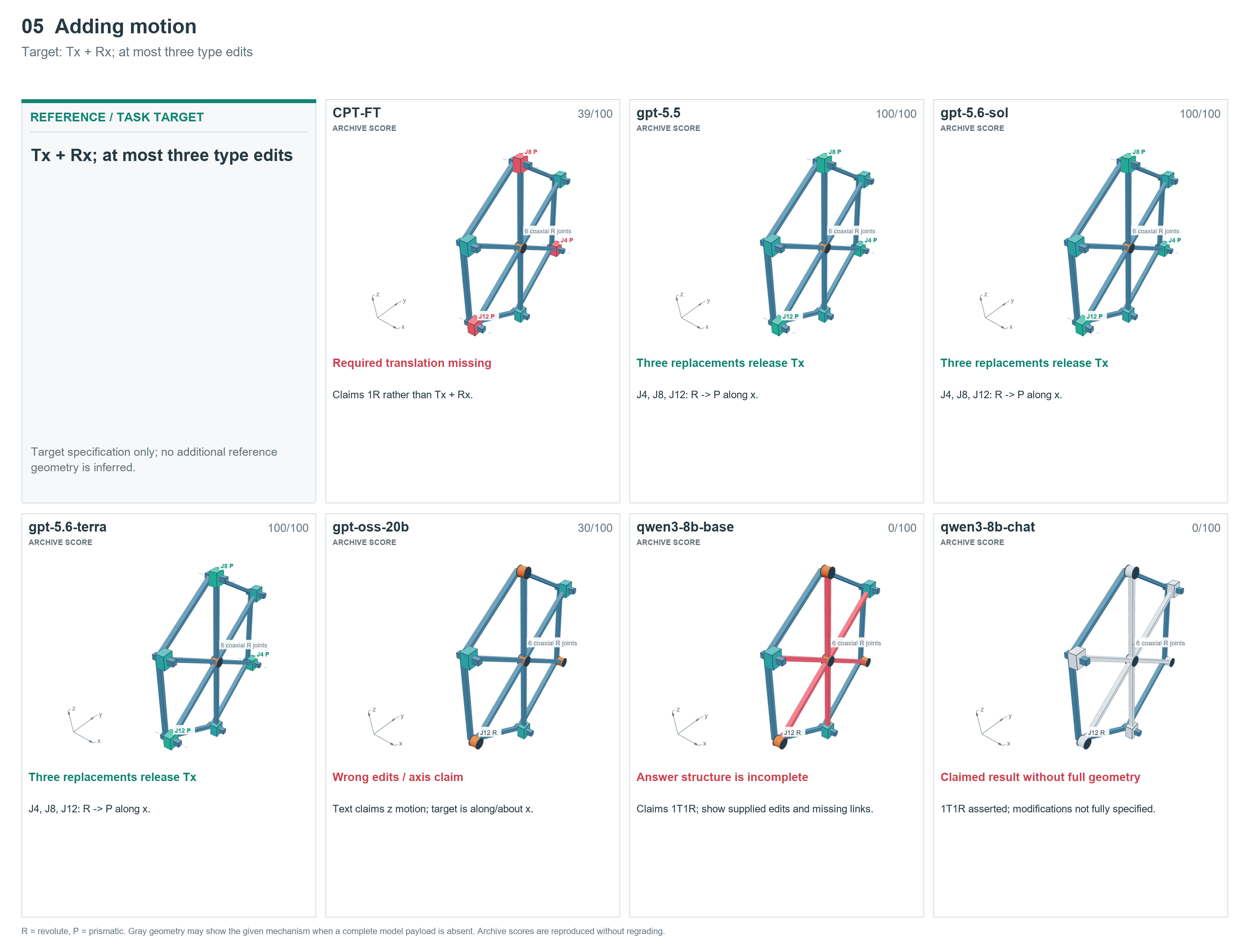}\\[4pt]
\includegraphics[width=0.90\textwidth,height=0.40\textheight,keepaspectratio]{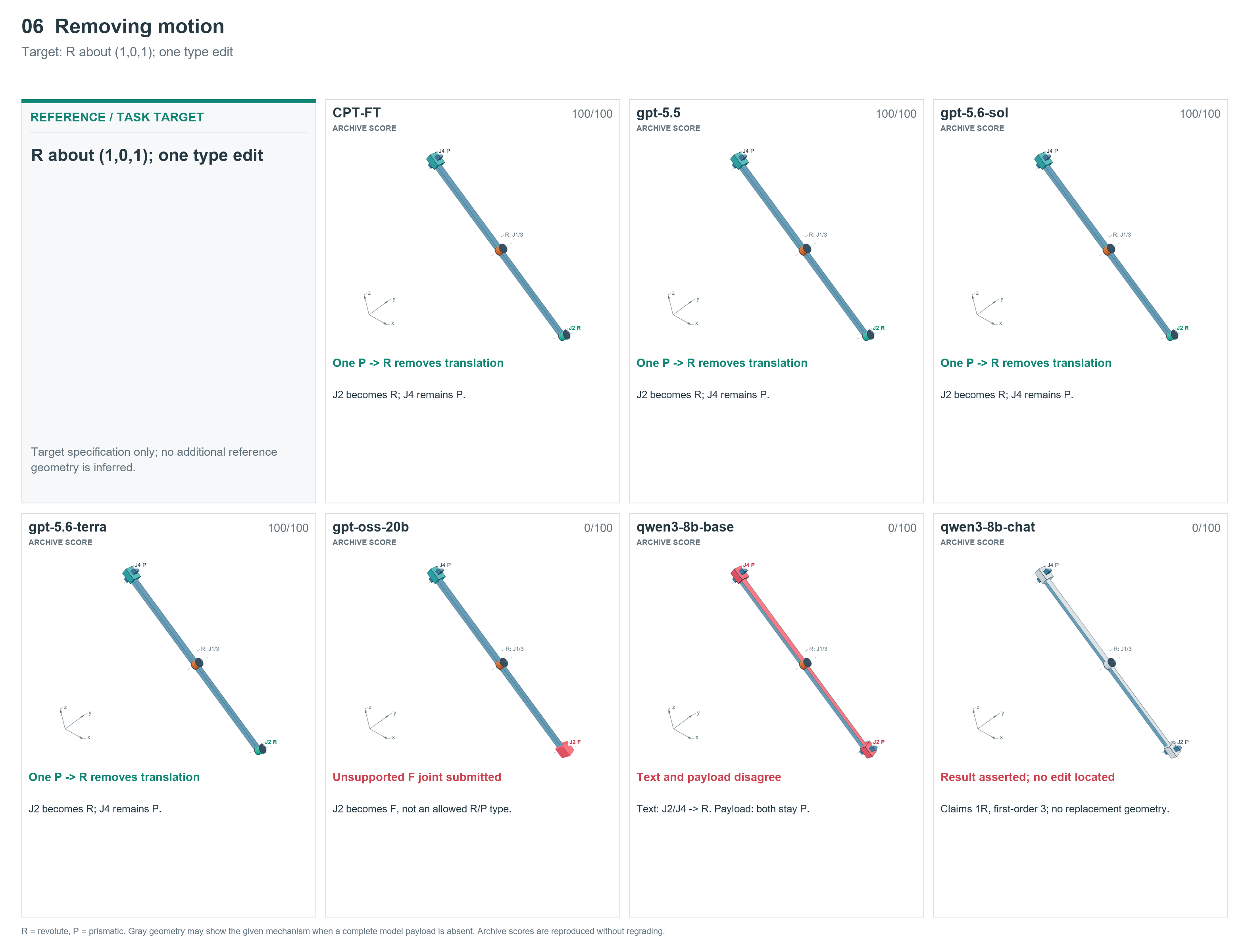}
\caption{Challenge response visualizations for categories 5--6.}
\end{figure}
\clearpage
\begin{figure}[H]
\centering
\includegraphics[width=0.90\textwidth,height=0.40\textheight,keepaspectratio]{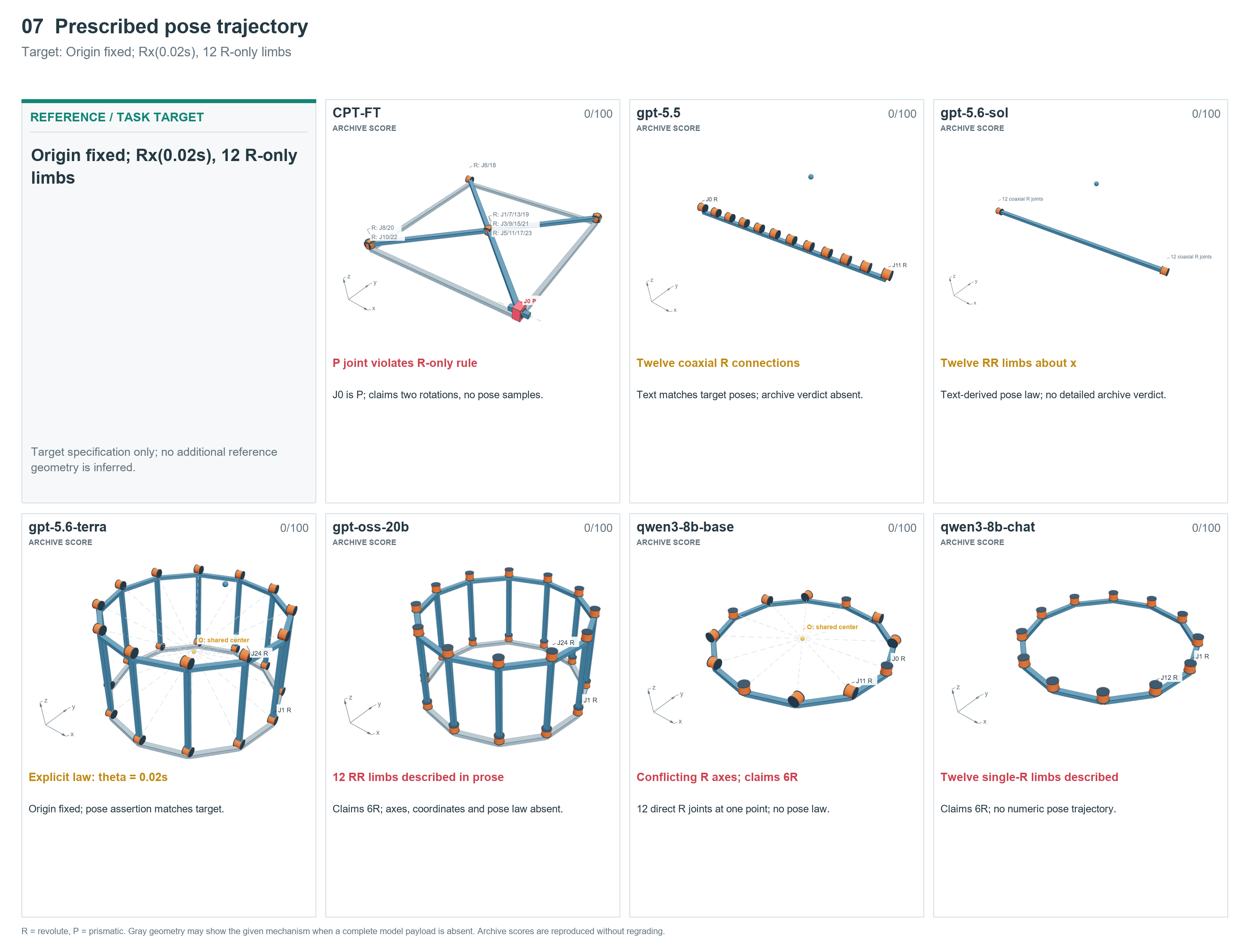}\\[4pt]
\includegraphics[width=0.90\textwidth,height=0.40\textheight,keepaspectratio]{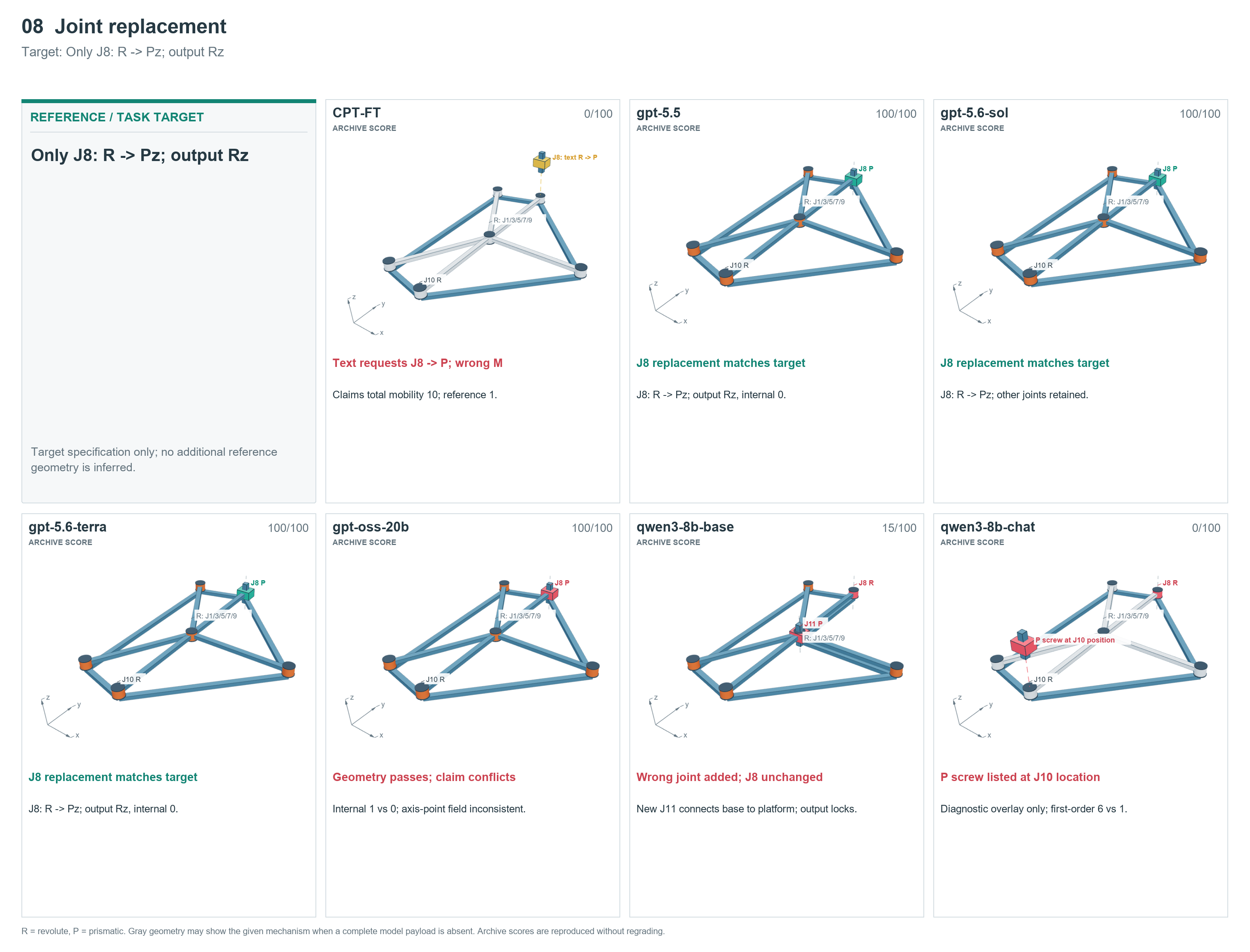}
\caption{Challenge response visualizations for categories 7--8.}
\end{figure}
\clearpage
\begin{figure}[H]
\centering
\includegraphics[width=0.90\textwidth,height=0.40\textheight,keepaspectratio]{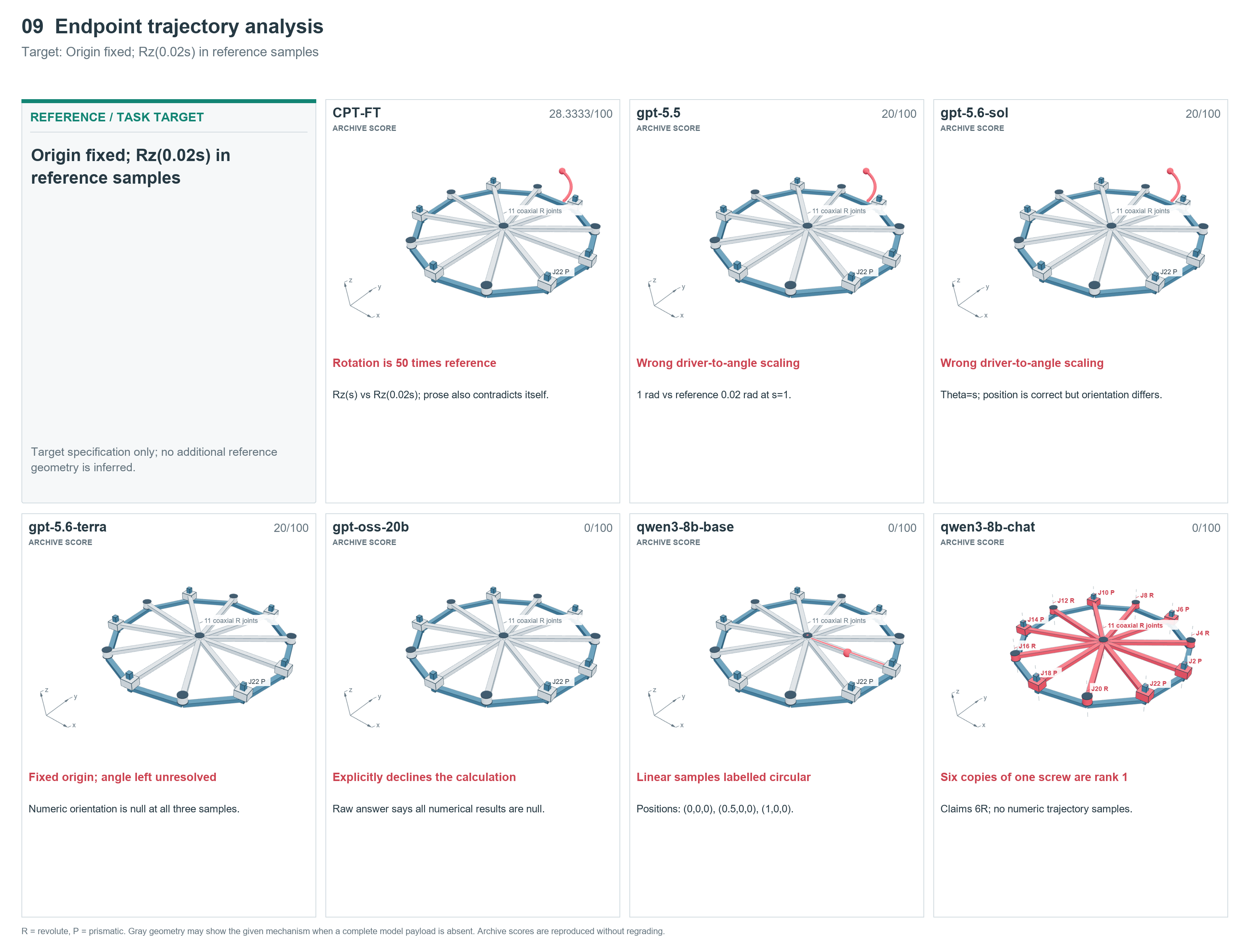}
\caption{Challenge response visualization, category 9.}
\label{fig:challenge_response_visualization}
\end{figure}

\clearpage
\section*{AI Use Statement}

Generative AI tools assisted candidate-question generation, model cross-checking, format conversion, and language editing. All candidate questions, reference answers, claims, citations, and AI-assisted content were reviewed by the authors or domain researchers. The authors remain responsible for the final manuscript and all AI-assisted artifacts.

\section*{Reproducibility Statement}

\revision{The 200 development questions are used only for protocol debugging and are excluded from reported benchmark results. Standard and challenge questions, reference answers, rubrics, extraction prompts, and program-verifier assets remain in the private server package.}

\section*{Ethics Statement}

The study uses professional-source-driven dataset construction and external model services. Data and code release will follow source licenses and service terms. \revision{Reference answers and item-level scoring resources remain under organizer control.} External judge services are used only after confirming data-use permissions and processing terms, and unverified model output is not treated as an engineering conclusion.

\begingroup
\expandafter\def\csname addedref@tian2020configurable\endcsname{1}
\expandafter\def\csname addedref@tian2024redundancy\endcsname{1}
\expandafter\def\csname addedref@xia2025gpmreview\endcsname{1}
\expandafter\def\csname addedref@xing2025kirigami\endcsname{1}
\expandafter\def\csname addedref@tian2017coupling\endcsname{1}
\expandafter\def\csname addedref@tian2021reconfigurable\endcsname{1}
\expandafter\def\csname addedref@xia2025transmission\endcsname{1}
\expandafter\def\csname addedref@xia2026mapping\endcsname{1}
\expandafter\def\csname addedref@tian2025review\endcsname{1}
\expandafter\def\csname addedref@gosselin1990singularity\endcsname{1}
\expandafter\def\csname addedref@mohamed2005configurable\endcsname{1}
\expandafter\def\csname addedref@lambert2016configurable\endcsname{1}
\expandafter\def\csname addedref@merlet2006\endcsname{1}
\let\originalbibitem\bibitem
\renewcommand{\bibitem}[2][]{%
  \color{black}%
  \originalbibitem[#1]{#2}%
  \ifcsname addedref@#2\endcsname\color{black}\fi
}
\bibliography{references}
\bibliographystyle{iclr2027_conference}
\endgroup

\end{document}